\documentclass{article}
\newcommand*{\affaddr}[1]{#1} % No op here. Customize it for different styles.
\newcommand*{\affmark}[1][*]{\textsuperscript{#1}}
\newcommand*{\email}[1]{\texttt{#1}}

\usepackage[preprint]{neurips_2019}

\usepackage[]{neurips_2019}

\usepackage{authblk}
\usepackage[utf8]{inputenc} % allow utf-8 input
\usepackage[T1]{fontenc}    % use 8-bit T1 fonts
\usepackage{hyperref} 
\usepackage{subfigure}
\usepackage{url}            % simple URL typesetting
\usepackage{booktabs}       % professional-quality tables
\usepackage{amsfonts}       % blackboard math symbols
\usepackage{amssymb}
\usepackage{nicefrac}       % compact symbols for 1/2, etc.
\usepackage[export]{adjustbox}
\usepackage{microtype}
\usepackage{float}% microtypography
\usepackage[]{natbib}
\usepackage{graphicx}
\usepackage{caption}
\usepackage{setspace}
\usepackage{multirow}
\usepackage{titlesec}
\titlespacing\section{0pt}{12pt plus 4pt minus 2pt}{0pt plus 2pt minus 2pt}
\titlespacing\subsection{0pt}{12pt plus 4pt minus 2pt}{0pt plus 2pt minus 2pt}
\titlespacing\subsubsection{0pt}{12pt plus 4pt minus 2pt}{0pt plus 2pt minus 2pt}
\graphicspath{ {NeuRIPS2019/} }
\begin{document}
\title{Fully Automated Image De-fencing using Conditional Generative Adversarial Networks}

% \author[1]{\small \textbf{Divyanshu Gupta} }
% \author[1]{\small \textbf{Shorya Jain}}
% \author[1]{\small \textbf{Utkarsh Tripathi}}
% \author[1]{\small \textbf{Pratik Chattopadhyay}}
% \author[2]{\small \textbf{Li-Po Wang}}

% \affil[1]{\footnotesize Department of Computer Science and Engineering, Indian Institute of Technology,(B.H.U), Varanasi, India}
% \affil[2]{\footnotesize Nanyang Technological University, Singapore}
% \email{\{A,B,C,D,E\}@university.edu}\\

\author{%
\textbf{Divyanshu Gupta}\affmark[1], \textbf{Shorya Jain}\affmark[1], \textbf{Utkarsh Tripathi}\affmark[1], \textbf{Pratik Chattopadhyay}\affmark[1], \textbf{Lipo Wang}\affmark[2]\\
\affaddr{\affmark[1]Department of Computer Science and Engineering, IIT(B.H.U), Varanasi-221005, India}\\
\affaddr{\affmark[2]School of Electrical and Electronic
Engineering, Nanyang Technological University, Singapore}\\ \vspace{0.2cm}
\email{\affmark[1]\{divyanshu.gupta.cse15,shorya.jain.cse16,utkarsh.tripathi.cse16\}@iitbhu.ac.in}\\%
\email{\affmark[1]pratik.cse@iitbhu.ac.in}\\%
\email{\affmark[2]elpwang@ntu.edu.sg}\\
}

\maketitle
\begin{abstract}
 Image de-fencing is one of the important aspects of recreational photography in which the objective is to remove the fence texture present in an image and generate an aesthetically pleasing version of the same image without the fence texture. In this paper, we aim to develop an automated and effective technique for fence removal and image reconstruction using %Motivated by the recent success of 
 conditional Generative Adversarial Networks (cGANs). These networks have been successfully applied in several domains of Computer Vision focusing on image generation and rendering. %, we study the applicability of these networks in our work. 
 Our initial approach is based on a two-stage architecture involving two cGANs that generate the fence mask and the inpainted image, respectively. Training of these networks is carried out independently and, during evaluation, the input image is passed through the two generators in succession to obtain the de-fenced image. The results obtained from this approach are satisfactory, but the response time is long since the image has to pass through two sets of convolution layers. To reduce the response time, we propose a second approach involving only a single cGAN architecture that is trained using the ground-truth of fenced de-fenced image pairs along with the %The model is first trained with a large set of fenced images and their corresponding de-fenced versions. Although the response time is reduced in this case,  the generated images are not always visually appealing, which might be due to the fewer number of layers compared to the previous architecture. To achieve better results, we propose to train the same network 
 edge map of the fenced image produced by the Canny Filter. %instead of training it with the pair of input image and de-fenced image alone. 
 Incorporation of the edge map helps the network to precisely detect the edges present in the input image, and also imparts it an ability to carry out high quality de-fencing in an efficient manner, even in the presence of a fewer number of layers as compared to the two-stage network. %, and subsequently the network to get trained to distinguish between fence and non-fence pixels accurately.
 Qualitative and quantitative experimental results reported in the manuscript reveal that the de-fenced images generated by the single-stage de-fencing network have similar visual quality to those produced by the two-stage network. Comparative performance analysis also emphasizes the effectiveness of our approach over state-of-the-art image de-fencing techniques.
\end{abstract}

\section{Introduction}
\label{sec:intro}
\begin{figure}[!ht]
    \centering
     \begin{subfigure}[]{ %{0.30\linewidth}
       \includegraphics[width=2.5cm,height=1.8cm,frame]{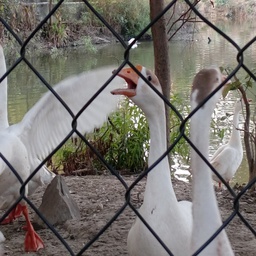}}
      \end{subfigure}
      \begin{subfigure}[]{ %{0.30\linewidth}
       \includegraphics[width=2.5cm,height=1.8cm,frame]{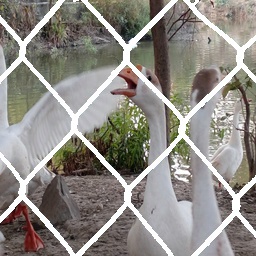}}
      \end{subfigure}
      \begin{subfigure}[]{ %{0.30\linewidth}
       \includegraphics[width=2.5cm,height=1.8cm,frame]{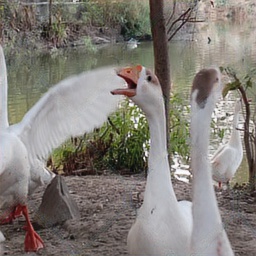}}
      \end{subfigure}
     \caption{(a) Input image with fence, (b) fence detection, (c) de-fenced image after inpainting.}
     \label{fig:fig1}
   \end{figure}

In recent times, due to the existence of inexpensive image capturing devices such as smart-phones and tablets with sophisticated cameras, there is a steady rise in the number of images captured and shared over the internet. Despite these technological advances, often it is difficult to take a snapshot of the intended object due to the presence of certain obstructions between the camera and the object. For example, consider that a person wishes to capture the image of an animal within a cage in a zoo. It is understandable that he/she will not be able to capture a clear image of the animal from a distance due to the presence of a fence or cage-bars in front. %Several computer vision based techniques have been developed to eliminate the fence texture from the image and generate a clear image of the scene. This specific task is known as image de-fencing.
%Image de-fencing refers to the task of generating an aesthetically pleasing image without the fence texture. It can be applied in several tasks dealing with occlusion removal so as to improve the performance of various computer vision algorithms such as object detection and recognition. It is easy to comprehend that, manual removal of the fence from the entire image is error-prone and also time-intensive. Hence, 
Till date, a number of research articles on image de-fencing have been proposed in the literature, (e.g., \cite{park2010image,khasare2013seeing,jonna2015mycamera,jonna2016deep,Liu2008cvpr, farid2016image}). However, these methods are either semi-automated, or are time-intensive due to involvement of complex computations. Usually, image de-fencing is viewed as a combination of two separate sub-problems: (i) \emph{fence mask generation}, which essentially clusters the image region into two groups, i.e., fence and non-fence regions, and (ii) \emph{image inpainting}, which deals with artificially synthesizing colors to the fence regions to make the rendered image look realistic. This is explained with the help of Figure \ref{fig:fig1}.
%Till date, a number of image de-fencing approaches have been developed, which have performed well with regular fence patterns (\cite{park2010image,khasare2013seeing,jonna2015mycamera,jonna2016deep,Liu2008cvpr}), but are not effective enough in the presence of irregular fence patterns in the image. Moreover, most of the work consider image de-fencing to be carried out in two important steps as depicted in Figure \ref{fig:fig1}: (a) fence detection, which essentially clusters the image region into two groups, i.e., fence and non-fence regions, and (b) inpainting, which deals with artificially synthesizing colours to the fence regions to make the rendered image looks realistic. 
%Although, till date, significant progress has been made in the area of automated and semi-automated fence detection,  %and recovery of the eliminated area to get fenced images. 
%automatic recovery of the plausible content behind the fence to generate a real looking de-fenced image is still a challenging problem.
 
Conditional Generative Adversarial Networks (cGANs) (\cite{reed2016generative,zhang2017stackgan,Huang2018AnIT}) have already demonstrated strong potential in generating realistic images in accordance with a set of user-defined conditions or constraints (\cite{yeh2017semantic,brkic2017know,radford2015unsupervised,chen2016infogan}), %a conditional generative model and make it 
and are possibly the best networks available today for performing any image-to-image translation task. Since image de-fencing can also be viewed as a sub-class of image-to-image translation problems, 
%Motivated by the huge success of Generative Neural Networks (GANs)  in the recent years in image generation and synthesis, e.g., \cite{yeh2017semantic,brkic2017know,brkic2017know,radford2015unsupervised,chen2016infogan}, 
we propose to use cGANs in our work as well. %GANs have already demonstrated strong potential in generating realistic images and have been applied in several areas including image synthesis  artificial colouring (\cite{yeh2017semantic}), face de-identifying (\cite{brkic2017know}), representation learning (\cite{radford2015unsupervised,chen2016infogan}), etc., and hence, we are applying GANs in this task.  Specifically, we use the Generative Adversarial Networks in a conditional setting, since, 

To the best of our knowledge, the present paper is the first work on image de-fencing that exploits the powerful generalization capability of GANs to generate de-fenced images. Specifically, we propose two different GAN-based de-fencing algorithms: (i) a two-stage network that consists of two sub-networks to carry out the fence mask generation and image inpainting in succession, and (ii) a single-stage network which directly performs image de-fencing without any intermediate fence mask generation step. Experimental results on a public fence segmentation data set (\cite{du2018accurateicme}) and our own artificially created data set  %\footnote{\url{https://drive.google.com/drive/folders/1CZqKTZX0TBBoAU7nSRJr1_X9n7G2x8BZ}} 
show the effectiveness of our work. %that the proposed GAN-based network capable of generating high quality de-fenced images in an efficient manner. %It has also been seen to perform better than other end-to-end image de-fencing techniques in the literature in terms of execution time and visual image quality In this work, we explore the applicability of Conditional Generative Adversarial Networks(cGANs), for Image de-fencing task, in two different settings, which removes the fence structures and occlusions from the single image. We experimented on the fence segmentation data set provided by \cite{du2018accurateicme}, as well as our own artificially created dataset, for both the settings, available online at \url{https://drive.google.com/drive/folders/1CZqKTZX0TBBoAU7nSRJr1_X9n7G2x8BZ}. The qualitative results confirm that our model removes the fences from a single image in an effective manner and performs better than the state-of-the-art techniques to a certain extent.

The rest of the paper is organized as follows: in Section \ref{sec:related_work}, we briefly review the related work in automatic fence detection, image inpainting and image de-fencing. The problem formalization and the proposed cGAN based de-fencing approaches are described in detail in Section \ref{sec:method}. An extensive experimental campaign is reported in Section \ref{sec:exp}. Finally, conclusions and future scopes for research are highlighted in Section \ref{sec:con}.

\section{Related Work}
\label{sec:related_work}
As explained before, traditionally, the task of image de-fencing is viewed as a two-step problem: (i) segmenting the fence in the image, and (ii) recovering the eliminated area with plausible content via image inpainting. Hence, in addition to reviewing the image de-fencing techniques in the literature, we also study existing solutions to fence mask generation and image inpainting, which are briefly described in the Sub-sections \ref{sec:fence_detection} and \ref{sec:image_inpain}, respectively. Finally, in Sub-section \ref{sec:image_def}, we present a review of the state-of-the-art image de-fencing techniques.

\subsection{Fence Mask Detection}
\label{sec:fence_detection}
Fence mask detection is the process of segmenting an input image with fence into two clusters, such that all the fence pixels are assigned a particular cluster, and each of the other pixels is assigned a different cluster. %(refer to Figures \ref{fig:fig1}(a) and (b)).
A lot of research has been done in the domain of regular and near-regular pattern detection (\cite{hays2006discovering,park2009deformed,lin2007lattice,park2010image}). The work in \cite{hays2006discovering} used higher-order feature matching to discover the lattices of near-regular patterns in real images based on the principal eigen vector of the affinity matrix. \cite{park2009deformed} proposed a method for detection of deformed 2D wallpaper patterns in real-world images by mapping the
2D lattice detection problem into a multi-target tracking problem, which can be solved within an Markov Random Field framework. In another work by \cite{park2008deformed}, the problem of near-regular fence detection was handled by employing efficient Mean-Shift Belief Propagation method to extract the underlying deformed lattice in the image.  \cite{mu2014video} proposed a soft fence-detection method using visual parallax as the cue to differentiate the fence from the non-fenced regions. 

%In the recent years, a number of deep learning based approaches have been used for automatic fence detection. For example,  \cite{jonna2015mycamera,jonna2016deep} uses convolutional neural networks for detecting fence pixels in a sequence of video frames, out of which \cite{jonna2015mycamera} first carries out depth completion by fusing multiple depth maps and next employs optical flow algorithm to find correspondences between frames. Finally, estimation of the de-fenced image is done by modeling it as a Markov Random Field, and obtaining its maximum a-posteriori estimate by applying loopy belief propagation. The work in  \cite{jonna2016deep} also follows a similar set of steps, the only difference being it uses fast iterative shrinkage thresholding algorithm (FISTA) to estimate the de-fenced image. %\cite{du2018accurateicme} proposed a fully convolutional network (FCN) (\cite{long2015fully}) based model to generate fence mask. 

\subsection{Image Inpainting}
\label{sec:image_inpain}
Image inpainting is the process of restoration of the unfilled portions of an image with appropriate plausible content/color. Image inpainting methods used in the literature can be broadly divided into two categories: (a) diffusion based methods (\cite{bertalmio2003simultaneous,levin2008closed}) and (b) exemplar-based methods (\cite{criminisi2004region,xu2010image,darabi2012image,huang2014image}). The former category of approaches uses smoothness priors to propagate information from known regions to the unknown region, while the latter category fills in the occluded regions by means of similar patches from other locations in the image. Exemplar-based methods have the potential of filling up large occluded regions and recreate missing textures to reconstruct large regions within an image. But these methods are unable to recover the high-frequency details of the image properly. To the best of our knowledge, Context Encoder (\cite{pathak2016context}) is the first deep learning approach used for image inpainting, in which an encoder is used to map an image with missing regions to a low-dimensional feature space, which is next used by the decoder to reconstruct the output image. \cite{yang2017high} used a pretrained VGG network that minimizes the feature differences in the image background, thereby improving the work of \cite{pathak2016context}. \cite{yeh2017semantic} proposed a GAN-based approach for image inpainting by applying a set of conditions on the available data. Contextual attention (\cite{Yu2018GenerativeII}) is another approach in which the missing regions were first estimated followed by an attention mechanism to sharpen the results. \cite{Nazeri2019EdgeConnectGI} developed a two-stage adversarial model consisting of an edge generator and image completion network. The edge generator network detects edges of missing regions and is used by the image completion network as prior to fill in the missing regions.
\subsection{Image De-fencing}
\label{sec:image_def}
The image de-fencing problem was first addressed in \cite{Liu2008cvpr}, where the fence patterns were segmented by means of spatial regularity, and the fence pixels were filled in with suitable colors by applying an appropriate inpainting algorithm (\cite{criminisi2004region}). \cite{park2010image} extended the work of \cite{Liu2008cvpr} by employing multiple images for extracting the information of occluded image data from additional frames. \cite{park2010image} also used a deformable lattice detection method similar to that of \cite{park2009deformed} discussed in Section \ref{sec:fence_detection}. \cite{khasare2013seeing} proposed an improved multi-frame de-fencing technique by using loopy-belief propagation (\cite{felzenszwalb2006efficient}). This method uses an image matting (\cite{zheng2009learning}) for fence segmentation, but the main drawback of this approach is that it involves significant user interaction and is therefore not very suitable for practical purposes. \cite{jonna2015multimodal} proposed a multimodal approach for image de-fencing in video frames, in which the fence mask was first extracted in each frame with the aid of depth maps corresponding to the color images obtained from a Kinect sensor, and next an optical flow algorithm was used to find correspondences between adjacent frames. Finally, estimation of the de-fenced image was done by modeling it as a Markov Random Field, and obtaining its maximum a-posteriori estimate by applying loopy belief propagation. \cite{kumar2016imagesignal} used signal demixing to capture the sparsity and regularity of the different image regions, thereby detecting fences, following which inpainting was performed to fill-in the fence pixels with suitable colors.
\cite{farid2016image} used a semi-automated approach in which an user is requested to manually mark several fence pixels in the image. A Bayesian classifier is next employed to classify each pixel as fence or non-fence pixel based on the knowledge of the color distribution of the marked pixels and the non-marked pixels. This approach is prone to human error and also highly time-intensive. %It further proposes a hybrid inpainting algorithm to generate the de-fenced image.

Recently, a few deep learning based video de-fencing approaches have been developed, e.g., the work by  \cite{jonna2015mycamera} is a semi-automated approach which first employs a CNN-based algorithm to detect the fence pixels in an input image, and next use a sparsity based optimization framework to fill-in the fence pixels. \cite{jonna2016deep} utilized a pre-trained CNN coupled with the SVM classifier for fence texel joint detection, and then connect the joints to obtain scribbles for image matting. %However, these deep learning based approaches suffer if fences with irregular patterns are provided. 
\cite{du2018accurateicme} presented an approach for fence segmentation using fully convolutional neural networks (FCN) (\cite{long2015fully}) and a fast robust recovery algorithm by employing occlusion-aware optical flow.

The main contributions of the paper are as follows:
\begin{enumerate}
\item Developing for the first time fully automated and efficient image de-fencing algorithms based on conditional Generative Adversarial Networks (cGANs).

\item Proposing a two-stage image de-fencing network that involves a fence mask generator and a image recovering network, each of which is based on cGAN. 
   
\item Making the algorithm more time-efficient by developing a single-stage end-to-end cGAN network without any intermediate fence mask generation step. Edge-based features along with the ground-truth of fenced de-fenced image pairs are used to train the model so that it can generate high quality de-fenced images even in the presence of a fewer number of layers compared to the two-stage network. 

\item Performing extensive experimental evaluation, and also making the codes and data set used in the experiments publicly available to the research community for further comparisons.
\end{enumerate}
\section{Proposed Techniques for Image De-fencing using cGANs}
\label{sec:method}
%We formulate the task of image de-fencing in two settings: 1) Two-stage image de-fencing, and 2) Single stage end-to-end image de-fencing. In former, both the stages follow an adversarial model (\cite{goodfellow2014generative}), i.e., each stage contains generator-discriminator pair. The latter contains a single generator-discriminator pair, along with some extra information regarding the fence structure in form of edge maps generated from Canny edge detector and Gabor filters. The main motive behind introducing single stage end-to-end image de-fencing network is due to high complexity of two-stage image de-fencing network, due to increased number of layers, since generator architecture is same for both the stages, only the objective function being different. However, for single stage end-to-end setting, the results were not visually pleasing (see Section \ref{sec:exp}). This could be due to absence of image priors, or some extra information that is required to detect fence structure. For tackling this issue, we used edge maps generated by Gabor filters and Canny filters as image priors, which improved the results. The complexity of single-stage network is reduced significantly as compared to two-stage network, and image quality is restored by using extra image priors in form of edge maps.  
The proposed cGAN-based two-stage and single-stage architectures for image de-fencing are explained in detail in the following two sub-sections. %Specifically, we use the Generative Adversarial Networks in a conditional setting, since, cGANs learn the conditional generative model and makes it suitable for various image-to-image translation tasks. We discuss the network architecture of both the settings in the following subsection.

\subsection{Two-stage Image De-fencing Network} \label{tsd}
\begin{figure}[h]
        \centering
        \includegraphics[width=14cm,height=4cm]{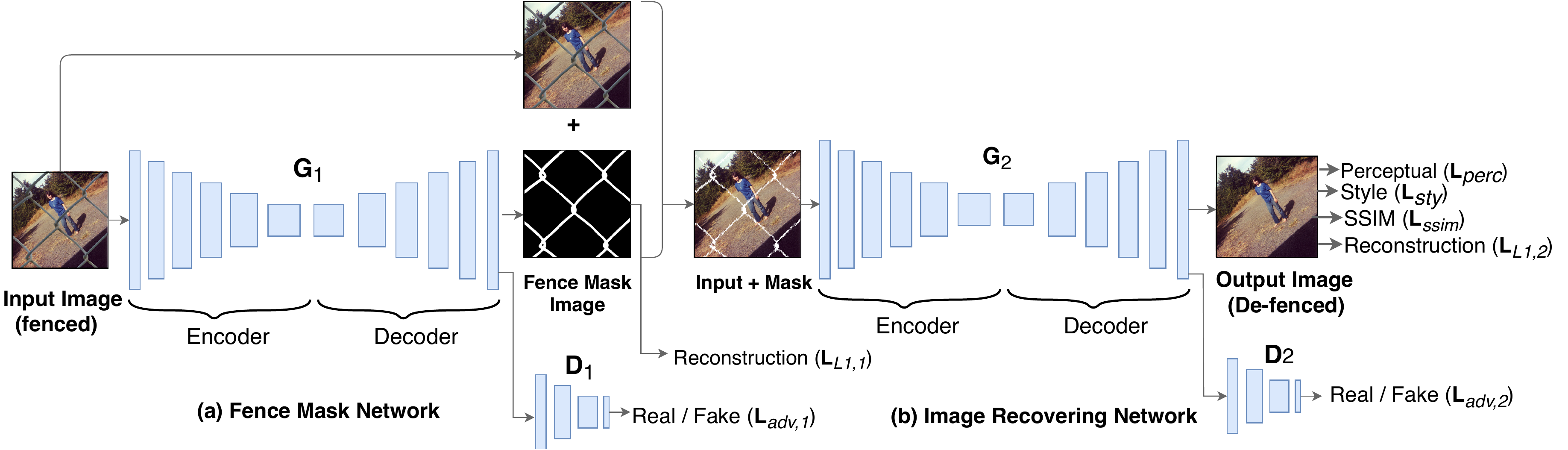}
        \caption{Two-stage image de-fencing network.}
        \label{fig:proposed}
\end{figure}

The two-stage architecture shown in Figure \ref{fig:proposed} follows a work-flow similar to that seen in most existing approaches, i.e., (i) fence mask generation and, (ii) recovering the missing parts of the image behind the fence (refer to Section \ref{sec:intro}). The difference from the existing techniques is that, we use adversarial learning to carry out both these steps. More specifically, we use a pair of cGANs, each consisting of a generator-discriminator pair, to perform the two steps. %The generator follows an architecture similar to \cite{isola2017image}, which has shown satisfactory results for applications related to image-to-image translation. 
The generator consists of a encoder network with seven down-sampling layers, followed by a decoder network with seven up-sampling layers as in \cite{isola2017image}. The discriminator is a $16{\times}16$ Markovian Discriminator, i.e, PatchGAN (\cite{isola2017image}), which classifies each $16{\times}16$ patch in the image as real or fake and averages all the responses to provide the final output. The proposed de-fencing algorithm is discussed in Sections \ref{sec:fmg} and \ref{sec:irn} by denoting ${D_1}$ and ${G_1}$ as the discriminator and generator for the fence mask generator, and ${D_2}$ and ${G_2}$ as the discriminator and generator for the image recovering network, respectively.

\subsubsection{Fence Mask Generator}
\label{sec:fmg}
The task of the fence mask generator can be viewed as an image-to-image translation problem where we convert a fenced image to a fence mask image. Let $\mathrm{I}_{mask}$ be the ground-truth fence mask image, and $\mathrm{I}_{fen}$ the input fenced image. During the training phase, the generator takes the input as $\mathrm{I}_{fen}$, conditioned on $\mathrm{I}_{mask}$ and predicts the fence mask image ${\mathrm{I}_{pred}}$. Mathematically, the generator function can be represented as:
\begin{equation}
    \mathrm{I}_{pred} = G_{1}(\mathrm{I}_{fen},\mathrm{I}_{mask}).
    \label{eq:1}
\end{equation}
We use ${I_{mask}}$ and ${I_{pred}}$, conditioned on ${I_{fen}}$ as the input of the discriminator, to predict the fence mask image as real or fake. The network is trained only with the objective function comprising adversarial loss and \emph{L1} loss, as shown next in (\ref{eq:2}):
\vspace{0.2cm}
\begin{equation}
    \min_{G_1}\max_{D_1} \mathrm{L}_{G_1} = \min_{G_1}( \alpha_{1} \max_{D_1}(\mathrm{L}_{adv,1}) ) + \beta_{1} (\mathrm{L}_{L1,1})), 
    \label{eq:2} 
\end{equation}
where $\alpha_{1}$ and $\beta_{1}$ are regularization parameters, with $\alpha_{1}=1$ and $\beta_{1}=10$. The adversarial loss ($\mathrm{L}_{adv,1}$) and \emph{L1} loss are defined in the following two equations: %(\ref{eq:3}) and (\ref{eq:4}), respectively, 
\vspace{0.2cm}
\begin{equation}
    \mathrm{L}_{adv,1} = \mathbb{E}_{(\mathrm{I}_{fen},\mathrm{I}_{mask})}[\log(D_{1}(\mathrm{I}_{mask},\mathrm{I}_{fen}))]
    + \mathbb{E}_{(\mathrm{I}_{fen},\mathrm{I}_{pred})}[\mathrm{log}(1- D_{1}(\mathrm{I}_{pred},\mathrm{I}_{fen}))],
    \label{eq:3}
\end{equation}
and
\begin{equation}
    \mathrm{L}_{L1,1} = \mathbb{E} [ \mathrm{\mid\mid} \mathrm{I}_{pred} -  \mathrm{I}_{mask} \mathrm{\mid\mid_{1}}], 
    \label{eq:4}
\end{equation}
where $\mathbb{E}$ denotes the expectation operator. The network is trained in multiple epochs, and training is stopped when the absolute difference of the value of loss function in two successive epochs is less than a small threshold $\epsilon$. We consider the value of $\epsilon$ as $10^{-3}$.
\subsubsection{Image Recovering Network}
\label{sec:irn}
\label{sec:img_rec}
Let ${\mathrm{I}_{def}}$ be the ground-truth de-fenced image. The input fenced image ${\mathrm{I}_{fen}}$ is masked with the fence mask image, which is obtained by the fence mask generator, say $\mathrm{M}$, to generate ${\mathrm{\tilde{\mathrm{I}}}_{fen}}$, i.e., ${\mathrm{\tilde{\mathrm{I}}}_{fen}}$ = ${\mathrm{I}_{fen}} {\odot} (1 - \mathrm{M})$ representing unfilled image without fences, where ${\odot}$ represents the Hadamard product. The image recovering network, takes ${\mathrm{\tilde{\mathrm{I}}}_{fen}}$ as input, conditioned on ${\mathrm{\mathrm{I}}_{def}}$ to generate a de-fenced image $\mathrm{\tilde{\mathrm{I}}}_{pred}$ filled with plausible content in the unfilled part of ${\mathrm{\tilde{\mathrm{I}}}_{fen}}$ having the same resolution as the input image, as shown in (\ref{eq:5}):
\begin{equation}
    {\mathrm{\tilde{\mathrm{I}}}_{pred}} = G_{2}({\mathrm{\tilde{\mathrm{I}}}_{fen}},\mathrm{I}_{def}).
    \label{eq:5}
\end{equation}

The image recovering network is trained until convergence by a joint objective function based on \cite{Nazeri2019EdgeConnectGI} that considers adversarial loss, perceptual loss, style loss and \emph{L1} loss, along with the SSIM loss (\cite{wang2004image}, \cite{zhao2017loss}). The adversarial loss function for $G_2$ is defined as:
\begin{equation}\label{eqob}
    \mathrm{L}_{adv,2} = \mathbb{E}_{(\tilde{\mathrm{I}}_{fen},\mathrm{I}_{def})}[\log(D_{2}(\mathrm{I}_{def},\tilde{\mathrm{I}}_{fen}))]
    + \mathbb{E}_{(\tilde{\mathrm{I}}_{fen},\tilde{\mathrm{I}}_{pred})}[\mathrm{log}(1- D_{2}(\tilde{\mathrm{I}}_{pred},\tilde{\mathrm{I}}_{fen}) )].
    \label{eq:6}
\end{equation}

The perceptual loss term (\cite{johnson2016perceptual}) $\mathrm{L}_{perc}$, computes the differences between the high-level feature representations between the ground-truth and the generated images extracted from a pre-trained CNN. If the predicted image label is dissimilar from the actual image label, a higher penalty is imposed through this loss term. Mathematically, the perceptual loss is defined as follows: %Equation (\ref{eq:7}).
\begin{equation}
    \mathrm{L}_{perc} = \mathbb{E}[ \sum_{i}^{} \frac{1}{N_i} \mid\mid \mathrm{\vec{a}_{i}}(\mathrm{I}_{def}) - \mathrm{\vec{a}_{i}}(\tilde{\mathrm{I}}_{pred}) \mid\mid_{1}],
    \label{eq:7}
\end{equation}
where $\mathrm{\vec{a}_{i}}$ is the activation map of the $i^{th}$ layer of a pre-trained VGG-19 network. We also use the style loss (\cite{gatys2016image}), which determines the difference between the style representations of two images. The style representation of an image at a particular layer  %is defined as the correlation between the filter responses, and 
is given by the gram matrix $\mathrm{G}$, and each element of this matrix %$\mathrm{G}^{l}$ at layer $l$%$\mathrm{G_{i,j}}^{l}$ is 
represents the inner product between a pair of  vectorized feature maps at the given layer. For the vectorized feature map of size $C_j \times H_j \times W_j$,  the style loss is mathematically defined as follows: %by Equation (\ref{eq:8}).
\begin{equation}
    \mathrm{L}_{sty} = \mathbb{E}_{j}[\mid\mid \mathrm{{G_{j}}^{\vec{a}}}(\tilde{\mathrm{I}}_{pred}) - \mathrm{{G_{j}}^{\vec{a}}}({\mathrm{I}}_{def}) \mid\mid_{1}],
    \label{eq:8}
 \end{equation}   
where $\mathrm{{G_{j}}^{\vec{a}}}$ is the a $\mathrm{C_j}\times \mathrm{C_j}$ 
gram matrix corresponding to feature map $\mathrm{\vec{a}_{j}}$. The \emph{L1} loss function is computed as follows: %defined in %by Equation (\ref{eq:9}).
\begin{equation}
    \mathrm{L}_{L1,2} = \mathbb{E} [ \mathrm{\mid\mid} \tilde{\mathrm{I}}_{pred} -  \mathrm{I}_{def} \mathrm{\mid\mid_{1}}  ].
    \label{eq:9}
\end{equation}
For obtaining visually pleasing images from the generator, we also incorporate a structural similarity loss term (\cite{wang2004image}, \cite{zhao2017loss}) as shown in (\ref{eq:9_2}), which indicates the differences in the luminance, contrast, and structure between the generated de-fenced image and the ground-truth de-fenced image. 
%SSIM for a pixel $p$ is defined by Equation :
\begin{equation}
    \mathrm{L}_{SSIM} =  \frac{1}{N} \sum_{p}^{}(1-\mathrm{SSIM(p)}),
    \label{eq:9_2}
\end{equation}
where, $p$ refers to a particular pixel position, and 
\begin{equation}\label{ssim}
    \mathrm{SSIM(p)} = \frac{2\mu_{x}\mu_{y} + C_1}{{\mu_{x}}^2 {\mu_{y}}^2 + C_1}.\frac{2\sigma_{xy} + C_2}{{\sigma_{x}}^2 {\sigma_{y}}^2 + C_2},
    \label{eq:9_1}
\end{equation}
refers to the structural similarity index magnitude between ground-truth and the generated image at pixel position $p$. In the above expression, $\mu_{x}$ and $\mu_{y}$ represent mean intensities in the neighborhood of $p$, while $\sigma_{x}$ and $\sigma_{y}$ represent standard deviations for two non-negative image signals $x$ and $y$, respectively, $\sigma_{xy}$ represents the co-variance of $x$ and $y$, and $C_1$ and $C_2$ are constants. Appropriate values for $C_1$ and $C_2$ to compute the SSIM loss can be found in \url{https://github.com/keras-team/keras-contrib/blob/master/keras_contrib/losses/dssim.py}.

The overall loss function for the image recovering network is henceforth computed as: %in (\ref{eq:10}).
\begin{equation}
    \min_{G_2}\max_{D_2} \mathrm{L}_{G_2} = \min_{G_2}( \alpha_{2} \max_{D_1}(\mathrm{L}_{adv,2})  + \beta_{2} (\mathrm{L}_{L1,2}) + \gamma (\mathrm{L}_{perc}) + \delta (\mathrm{L}_{sty}) + \eta (\mathrm{L}_{SSIM}) ),
    \label{eq:10} 
\end{equation}
where $\alpha_{2}$, $\beta_{2}$, $\gamma$,  $\delta$ and $\eta$ are regularization parameters. In our experiments, we use $\alpha_{2}$=0.1, $\beta_{2}$=10, $\gamma$=2, $\delta$=1 and $\eta$=1.

\subsection{Single Stage End-to-End Image De-fencing Network}\label{ssd}
\begin{figure}[h]
        \centering
        \includegraphics[width=12cm,height=4cm]{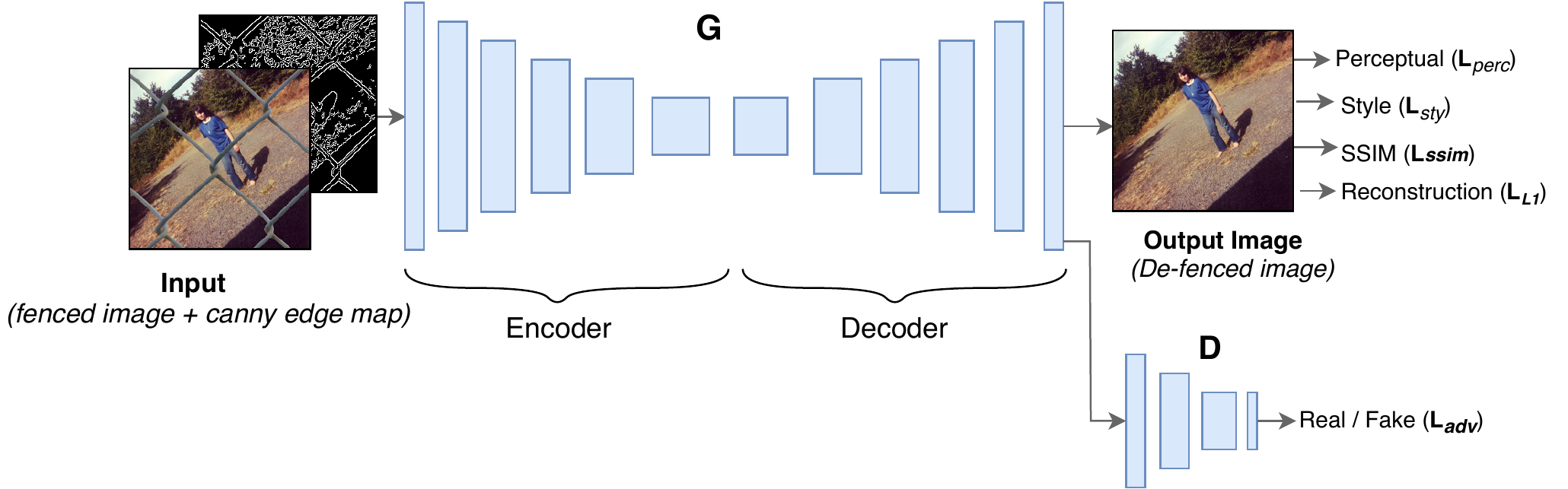}
        \caption{Single-stage image de-fencing network.}
        \label{fig:proposed2}
\end{figure}
Use of two generators makes the image de-fencing process time-intensive. To reduce the response time, we propose to use a single end-to-end architecture %as shown in The proposed architecture, in %. is a single stage end-to-end image de-fencing network architecture, which 
with a single generator-discriminator pair for translating the input image with a fence structure directly to its de-fenced version without any fence generation step in between. Since only a single generator is used here, the single-stage network will have fewer number of layers compared to the two-stage network described in Section \ref{tsd}. Presence of fewer layers in a convolutional neural network has a tendency to loose significant contextual information. To boost up the performance of the single-stage network, we propose to train the model %to predict the de-fenced image directly from 
with the given ground-truth fenced de-fenced image pairs as well as with an edge map of the fenced image given by the Canny filter. We observe that appending the Canny edge map along with the input fenced image facilitates high quality image de-fencing even in the presence of fewer number of layers.
%The fenced image is supplied along with some extra information regarding the fence structure in the form of edge maps generated from Canny edge detector. % and Gabor filters. We used Gabor filters (\cite{mehrotra1992gabor}), since when it is applied to an image, it gives the maximum response at the points and edges where the fence texture changes, therefore, the filter has a distinguishing value at the spatial location of the feature, for which it gives maximum response. The notable parameters of Gabor filter includes kernel size, orientation of the normal to parallel stripes of Gabor function, wavelength of sinusoidal factor etc. We also experimented with Canny filter for edge detection (\cite{canny1987computational}), due to its great ability to extract fence edge maps, by showing the positions of tracked intensity discontinuities. It is based on calculating gradient of the image and a hysteresis filtering, which enables to select the lines of adjacent pixels, contrasting with their neighbors. The output of Gabor filter and Canny filter for edge detection for a given fenced image are depicted in Figure \ref{fig:cannygabor}.
%\begin{figure}[h]
%        \centering
%        \includegraphics[width=9cm,height=2.8cm]{gaborcanny.jpg}
%        \caption{Output of Gabor and Canny filters on a given %fenced image}
%        \label{fig:cannygabor}
%\end{figure} 
The network architectures for the generator and discriminator are similar to that of the image recovering network in Section \ref{sec:img_rec}. Also, an objective function similar to (\ref{eq:10}) has been used to train the single-stage network until convergence. %is similar to that specified in Equation (\ref{eq:10}). 
%The only modification is the usage of extra edge information in the form of Edge maps, which is concatenated with the input fenced image during the training phase. The extra edge information generated by Canny edge detector and Gabor filter, help to identify the fence structure and enable the network to learn to remove the fence structure.
\section{Experiments and Results} \label{sec:exp}
\subsection{Data Set and System Description}
We use three Graphics Processing Units (GPUs) to train the models out of which one is Nvidia Titan Xp (with 12GB RAM, total FB memory as 12196 MB and total BAR1 memory as 256 MB), and the other two are Nvidia GeForce GTX 1080 Ti (with 11 GB RAM, total FB memory as 11178 MB and total BAR1 memory as 256 MB). 
%\subsection{Dataset Description}
The experimental protocols are briefly discussed next. %We evaluated our proposed approach for image de-fencing in both the settings on fence dataset described as follows: 
For training the two-stage image de-fencing network, the fence segmentation dataset provided by \cite{du2018accurateicme} is used along with a synthetically generated data set formed by adding artificial fence structures on a set of images from the Pascal VOC dataset (\cite{everingham2010pascal}). For training the image recovering network, we also created an artificial data set by using images from the Pascal VOC dataset (\cite{everingham2010pascal}) and the COCO dataset (\cite{lin2014microsoft}). The data set for training the single-stage end-to-end image de-fencing network is also constructed in a similar manner %, we also created the artificial dataset, in the same way as for image recovering network, 
by applying random fence structures on a set of images from Pascal VOC data %(\cite{everingham2010pascal}) 
and COCO data. The test set consists of a total of 245 images and is formed by selecting images from the above-mentioned public data sets as well as some images captured by our research team. Before, making a forward pass through the network, the dimensions of each training and test image is made equal to 256$\times$256. %The synthetically generated data sets for training the two networks, the test set, as well as the source code are made available in the following link: \url{https://drive.google.com/drive/folders/1CZqKTZX0TBBoAU7nSRJr1_X9n7G2x8BZ}.  
%(\cite{lin2014microsoft}).
%The qualitative and quantitative results of our experimentation in both the settings discussed earlier are shown in following section-

\subsection{Experiments with Two-Stage Image De-fencing Network}\label{resultssec}
\begin{figure}[h]
        \centering
        \includegraphics[width=13.5cm,height=5.0cm]{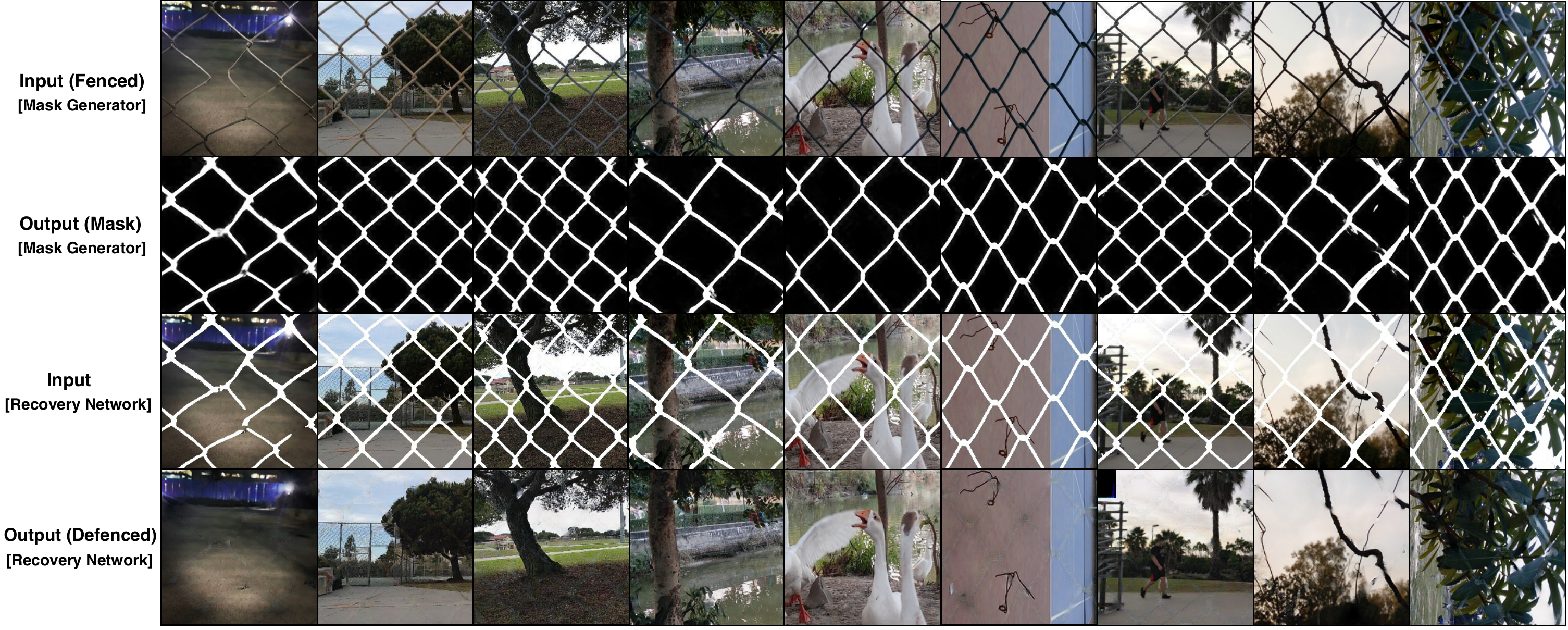}
        \caption{Qualitative results for the two-stage image de-fencing network.}
        \label{fig:exp1}
\end{figure}

Figure \ref{fig:exp1} shows the qualitative performance of the two-stage image de-fencing network by means of a sample set of fenced images. The first row in the figure represents the input, the second row represents the output of the mask generator, the third row corresponds to the input to the image recovery network, and finally the last row shows the output of the image recovery network.
It is seen from the figure that the output results are visually quite appealing. %But to critically examine our approach, we compute the MOS scores for the images generated by the two-stage network as well as its average response time after processing all the images in the test set. Table \ref{t1} presents the average response time of the fence mask generator and the image recovery network  separately and also the expected response time of the complete de-fencing network consisting of the above two generators. Table shows that the average response time of the two-stage network is 0.460 seconds. % after executing the algorithm on all the test images.

\subsection{Experiments with Single-Stage Image De-fencing Network}
\begin{figure}[h]
        \centering
        \includegraphics[width=13.5cm,height=6cm]{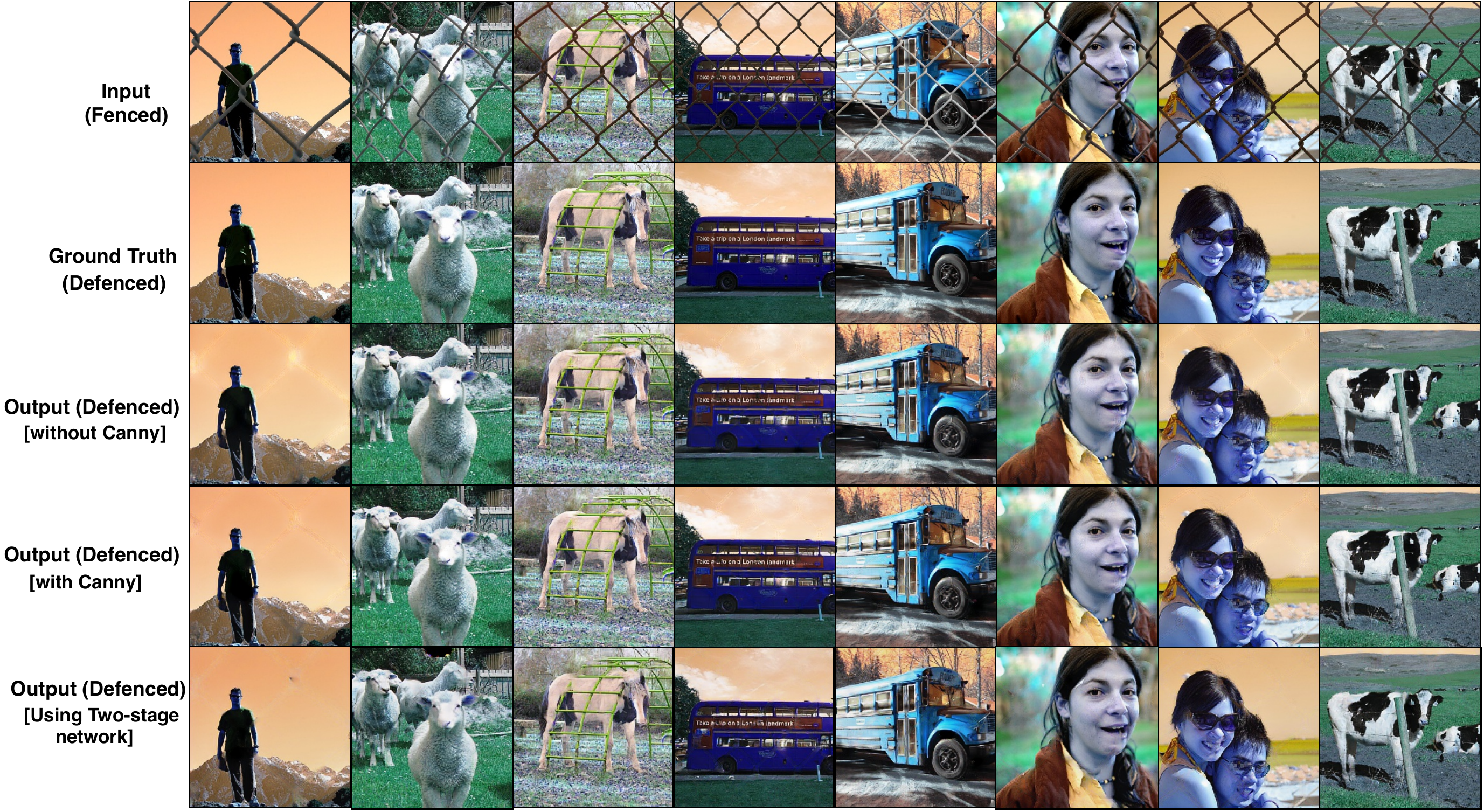}
        \caption{Qualitative results by the single stage end-to-end image de-fencing network.}
        \label{fig:exp2}
\end{figure}

As discussed in Section \ref{ssd}, in a bid to reduce the response time of image de-fencing further, we propose an efficient single-stage image de-fencing network. %is trained with the original image along with its Canny edge map. 
In the fourth row of Figure \ref{fig:exp2}, the outputs given by the proposed single-stage de-fencing network (with supervision of canny edge map) are shown on eight different test images. The first and second rows of the figure respectively depict the input image and the corresponding ground-truth, %The outputs of the generator trained with only the ground-truth fenced de-fenced image pairs (and without any auxiliary edge map information) are shown in the fourth row of the same figure. 
while the third row shows the images generated by single-stage image de-fencing network without extra supervision of canny edge maps, fifth row shows the images generated by the two-stage de-fencing network discussed in Section \ref{tsd}.
%addition to the input image during training the model, we also show
%We performed extensive experiments on the single-stage end-to-end image de-fencing network in two variants: one without extra edge information, and another with extra edge information.
%\subsubsection{Without extra edge information}
The qualitative results clearly depict that the output images given by the proposed single stage image de-fencing network %without auxiliary information are not visually pleasing and the de-fencing is not perfect. On the other hand, the proposed single-stage de-fencing network with auxiliary information by means of Canny edge map is able to retain the precision and quality of the output image similar to 
are comparable with that of the two-stage de-fencing network. %To further verify the effectiveness of appending the edge map during training the single-stage de-fencing network, in Table \ref{t2} we present a comparative performance analysis of this network by training it two times with and without the Canny edge map information. Results are shown in the table by means of SSIM score (refer to (\ref{ssim})) and average MOS. Corresponding results for the two-stage network are also shown in the same table. 
\section{Conclusions and Future Work}\label{sec:con}
In the present manuscript, we explore the applicability of conditional Generative Adversarial Networks (cGANs) for image de-fencing in two different settings: (i) a two-stage image de-fencing network which employs a fence mask generator network and an image recovering network in succession, and (ii) a single-stage image de-fencing network, which uses a single generator to render the de-fenced image from the original fenced image and its edge map. Experimental results show that both the proposed methods can produce de-fenced images with high visual quality, but the later approach is more advantageous if a fast response is desired. Comparison with state-of-the-art techniques also emphasize the effectiveness of employing GANs in the image de-fencing task. In future, it may be studied how to make the algorithm more time-efficient, so that it can be applied for real-time video de-fencing. %As seen from Table \ref{t2}, the average response time of the proposed single-stage image de-fencing network is 0.188 seconds, which implies that it can process about 5.26 frames per second. Thus, the same single-stage de-fencing network can conveniently handle a 5 frames/second video, but it is not efficient enough to process videos with high frame-rate. 
Another challenging area of research is how to effectively de-fence images with irregular or repeated fence structures, which the present work and also other existing techniques are unable to handle properly. %Modifying the training set . %None of the automated de-fencing approaches is capable of handling these challenging situations effectively. 
%GAN architecture, along with extra edge information about fences with the help of Canny edge detector and Gabor filters, and generates a de-fenced image from given fenced image. We experimentally show the results on datasets to confirm that our model removes the fences from a single image in an effective manner.
%\begin{figure}[h]
%        \centering
%        \includegraphics[width=10cm,height=4cm]{un%seenirregular_future.png}
%        \caption{Results of proposed de-fencing 
%networks on irregular and sparse fence structures}
%        \label{fig:irreg}
%\end{figure}
%The first column of Figure \ref{fig:irreg} shows two such cases with sparse and repetitive fence structures, respectively, while the fourth and fifth columns show the rendered images produced by the two-stage and single-stage networks, respectively. It can be clearly seen from the figure that the results fail to match the expectation. Similar performances were observed for other state-of-the-art de-fencing techniques as well. Enhancing the gallery set by adding synthetically generated repeated fence structures might be able to solve this problem.
\bibliographystyle{plainnat}  
\bibliography{neurips_2019}
%/
% \small
% [1] Alexander, J.A.\ \& Mozer, M.C.\ (1995) Template-based algorithms for
% connectionist rule extraction. In G.\ Tesauro, D.S.\ Touretzky and T.K.\ Leen
% (eds.), {\it Advances in Neural Information Processing Systems 7},
% pp.\ 609--616. Cambridge, MA: MIT Press.

% [2] Bower, J.M.\ \& Beeman, D.\ (1995) {\it The Book of GENESIS: Exploring
%   Realistic Neural Models with the GEneral NEural SImulation System.}  New York:
% TELOS/Springer--Verlag.

% [3] Hasselmo, M.E., Schnell, E.\ \& Barkai, E.\ (1995) Dynamics of learning and
% recall at excitatory recurrent synapses and cholinergic modulation in rat
% hippocampal region CA3. {\it Journal of Neuroscience} {\bf 15}(7):5249-5262.
\end{document}